\documentclass[letterpaper]{article} 
\usepackage{aaai2026}  
\usepackage{times}  
\usepackage{helvet}  
\usepackage{courier}  
\usepackage[hyphens]{url}  
\usepackage{graphicx} 
\usepackage{natbib}  
\usepackage{caption} 
\usepackage{algorithm}
\usepackage{algorithmic}
\usepackage{paralist}

\usepackage{newfloat}
\usepackage{listings}
\DeclareCaptionStyle{ruled}{labelfont=normalfont,labelsep=colon,strut=off} 
\floatstyle{ruled}
\newfloat{listing}{tb}{lst}{}
\floatname{listing}{Listing}
\usepackage{booktabs}
\usepackage{multirow}
\usepackage{amsfonts}
\usepackage{amsmath}
\usepackage{dsfont}
\usepackage{xcolor}

\title{AD-FM: Multimodal LLMs for Anomaly Detection via Multi-Stage Reasoning and Fine-Grained Reward Optimization}
\author{
    \textbf{Jingyi Liao\textsuperscript{\rm 1,\rm 2}, Yongyi Su\textsuperscript{\rm 2,\rm 3}, Rong-Cheng Tu\textsuperscript{\rm 1*}, Zhao Jin\textsuperscript{\rm 1}, Wenhao Sun\textsuperscript{\rm 1}, Yiting Li\textsuperscript{\rm 2}, Xun Xu\textsuperscript{\rm 2}\thanks{Corresponding authors}, \\ Dacheng Tao\textsuperscript{\rm 1}, Xulei Yang\textsuperscript{\rm 2}}\\
}
\affiliations{ 
    \textsuperscript{\rm 1}Nanyang Technological University \\
    \textsuperscript{\rm 2}Institute for Infocomm Research (I$^2$R), A*STAR \\
    \textsuperscript{\rm 3}South China University of Technology
}

\begin{document}

\maketitle

\begin{abstract}
While Multimodal Large Language Models (MLLMs) demonstrate remarkable capabilities across diverse domains, their application to specialized anomaly detection (AD) remains constrained by domain adaptation challenges. Existing Group Relative Policy Optimization (GRPO) based approaches suffer from two critical limitations: inadequate training data utilization when models produce uniform responses, and insufficient supervision over reasoning processes that encourage immediate binary decisions without deliberative analysis.
We propose a comprehensive framework addressing these limitations through two synergistic innovations. First, we introduce a multi-stage deliberative reasoning process that guides models from region identification to focused examination, generating diverse response patterns essential for GRPO optimization while enabling structured supervision over analytical workflows. Second, we develop a fine-grained reward mechanism incorporating classification accuracy and localization supervision, transforming binary feedback into continuous signals that distinguish genuine analytical insight from spurious correctness.
Comprehensive evaluation across multiple industrial datasets shows that our method achieves superior accuracy by enabling general-purpose MLLMs to acquire fine-grained visual discrimination for detecting subtle manufacturing defects. 
\end{abstract}


\section{Introduction}

Anomaly detection typically involves modeling the distribution of normal data to identify deviations. Recent advances in multi-modal learning have enabled zero-shot generalization by incorporating language descriptions~\cite{winclip,anomalyclip}. The emergence of multimodal large language models (MLLMs) further expands this capability by allowing natural language reasoning over visual inputs~\cite{jiang2025mmadcomprehensivebenchmarkmultimodal,tu2025mllm, gu2023anomalygptdetectingindustrialanomalies}.

\begin{figure}[!htbp]
    \centering
    \includegraphics[width=\linewidth]{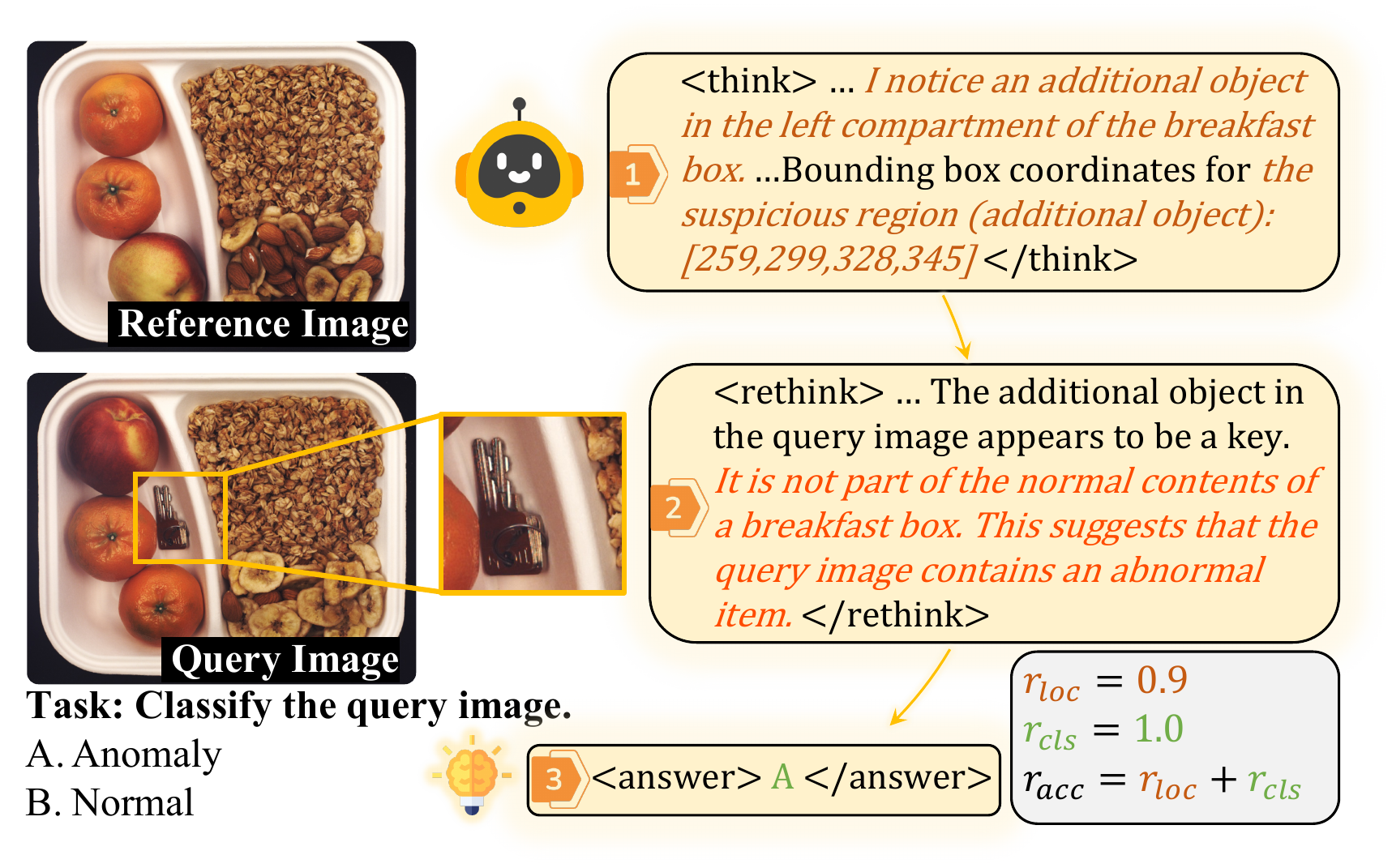}
    \caption{\footnotesize{Deliberative inference pipeline of AD-FM. The model performs multi-stage reasoning by identifying suspicious regions (\texttt{<think>}), focused region analysis (\texttt{<rethink>}), and making a final decision (\texttt{<answer>}). The localization reward $r_{\text{loc}}$ and classification reward $r_{\text{cls}}$ jointly form the fine-grained reward.}
}
    \label{fig:pipeline}
\end{figure}

However, recent studies~\cite{jiang2025mmadcomprehensivebenchmarkmultimodal,ader} show that general-purpose MLLMs struggle with specialized tasks like industrial anomaly detection, where fine-grained visual cues, e.g. subtle surface defects require reasoning beyond what is captured in their pretraining. This challenge motivates a paradigm of adapting general MLLMs towards downstream anomaly detection tasks.

Reinforcement fine-tuning (RFT)~\cite{ouyang2022training} offers a promising adaptation path by enabling task-specific reasoning without needing additional supervision. Compared to supervised fine-tuning (SFT), RFT facilitates greater exploration and generalization by optimizing against a reward signal derived from a simple reward function, thereby circumventing the reliance on detailed, token-level reasoning labels.
Notably, Group Relative Policy Optimization (GRPO)~\cite{shao2024deepseekmath} has shown potential when applied to MLLMs for anomaly detection~\cite{AnomalyR1,omniAD}. Yet, existing applications of GRPO to anomaly detection suffer from two critical limitations. First, GRPO's fundamental reliance on response/reward variance for learning signals leads to severely reduced data utilization efficiency when the model produces uniform outputs. Second, the lack of supervision over the reasoning process allows flawed analytical patterns to persist unchecked.

In GRPO, multiple responses to a prompt are scored and normalized (e.g., via z-score). When responses are uniform, e.g. consistently incorrect/correct, the variance collapses, producing vanishing gradients and halting learning. This is particularly problematic in downstream anomaly-detection tasks under domain shift, where the binary nature of classification decisions creates an inherently constrained reward space that severely limits response diversity. Moreover, even consistently correct answers may rely on flawed reasoning due to insufficient supervision of the thinking process. For instance, a model may detect an anomaly but reference an irrelevant region, or fail to identify all anomalous regions, which we term as \textit{spurious correctness}. 

To address these challenges, we introduce two synergistic improvements: a multi-stage deliberative reasoning framework and fine-grained reward mechanisms.

We first restructure the reasoning process to encourage more careful and methodical visual analysis, which is then utilized for a more fine-grained reward signal.
Motivated by recent successes in test-time scaling and chain-of-thought prompting~\cite{madaan2023self,kojima2022large, wei2022chain}, and adding additional reasoning process in GRPO~\cite{grit}, we propose a multi-stage deliberative reasoning framework that encourages more cautious decision-making while creating new supervisable intermediate signals. As illustrated in Fig. \ref{fig:pipeline}, our method guides models through a structured process: first identifying and localizing regions of interest, then performing focused analysis within these regions before producing final decisions. This approach aligns with human expert cognitive strategies during visual inspection, i.e., rapid global scanning followed by focused examination of suspicious areas, while improving traceability and reliability of the model’s decision process.

With models now producing detailed spatial predictions, we introduce complementary reward mechanisms that leverage this enriched information. These rewards deliver dual benefits: enhanced reward variance for improved GRPO learning and systematic filtering of spurious correctness patterns.
Inspired by multi-objective GRPO frameworks~\cite{li2025optimizing}, we design fine-grained reward functions that assess both decision accuracy and localization quality. Specifically, we prompt the model to not only provide a decision and rationale but also localize the anomalies via bounding boxes. Since defects may spread out on the object of interest, our first reward measures count accuracy of predicted boxes against ground truth, while the second computes Generalized Intersection over Union (GIoU) between predicted and ground-truth boxes. These spatially grounded rewards enable fine-grained discrimination between genuine understanding and spurious correctness, allowing models to exploit available data more effectively while developing reliable visual analysis capabilities, 
thereby enhancing deployment safety and trustworthiness in inspection scenarios.

The synergy between multi-stage reasoning and fine-grained rewards creates a comprehensive framework that simultaneously addresses GRPO's variance dependency and the challenge of supervision for reasoning process, transforming both the learning efficiency and reasoning quality of GRPO-based anomaly detection systems. In summary, we propose a framework for \textbf{A}nomaly \textbf{D}etection via \textbf{F}ine-grained reward optimization and \textbf{M}ulti-stage reasoning (\textbf{AD-FM}). Our key contributions are as follows:

\begin{itemize}
    \item We introduce a multi-stage deliberative reasoning framework that creates supervisory signals for reasoning and enables systematic visual inspection for AD.
    \item We design fine-grained rewards that suppress spurious correctness and maximize data utilization via joint optimization of classification accuracy and spatial understanding.
    \item We demonstrate substantial performance improvements across multiple datasets in both in-distribution and out-of-distribution settings, effectively bridging the gap between general-purpose MLLMs and specialized visual inspection requirements.
\end{itemize}

\section{Related Works}
\label{gen_inst}

\noindent\textbf{Anomaly Detection.}
Anomaly Detection (AD) involves identifying instances that deviate from normal patterns. Modern AD techniques are predominantly unsupervised and aim to model the distribution of normal data \cite{ader}. Key approaches include: one-class classification \cite{liu2023simplenet, liao2024coft}, reconstruction-based methods that identify anomalies via reconstruction error \cite{he2024mambaad, zhang2024realnet, li2025find}, feature distillation which compares original and distilled features \cite{deng2022anomalydetectionreversedistillation}, and embedding-based strategies that measure feature divergence between normal and target samples \cite{roth2022totalrecallindustrialanomaly, Lei_2023_CVPR}.

\noindent\textbf{Zero-shot Anomaly Detection.}
Zero-shot AD leverages textual prompts to detect anomalies in unseen distributions, enabling deployment without retraining. Early work used CLIP-based models \cite{li2025myriadlargemultimodalmodel, gu2023anomalygptdetectingindustrialanomalies} to align visual features with anomaly-related text, while \cite{segano} incorporated SAM \cite{kirillov2023segment} for improved localization. The rise of Multimodal LLMs (MLLMs) \cite{openai2024chatgpt, hurst2024gpt-4o} expanded their use in vision tasks, including anomaly detection \cite{cao2023genericanomalydetectionunderstanding, deng2024vmadvisualenhancedmultimodallarge}. Although early attempts \cite{li2025myriadlargemultimodalmodel, gu2023anomalygptdetectingindustrialanomalies} explored few-shot prompting, they lacked binary decision capability. MMAD \cite{jiang2025mmadcomprehensivebenchmarkmultimodal} introduced the first MLLM-based binary AD benchmark, but both MMAD and \cite{chen2025multimodallargelanguagemodels} reported limited zero/one-shot performance despite improved explainability.

\noindent\textbf{Reinforcement Fine-tuning.}
Reinforcement fine-tuning (RFT), first applied in O1 \cite{jaech2024openai} and DeepSeek R1 \cite{guo2025deepseek}, has proven effective for structured reasoning tasks. Multimodal RFT \cite{liu2025visual, chen2025r1v} extends this to vision with task-specific reward functions, while \cite{grit} introduces a dual-thinking process but lacks supervision over intermediate steps. However, applying RFT to AD remains nascent. AnomalyR1 \cite{AnomalyR1} used GRPO directly without addressing AD-specific issues such as sparse binary rewards and low feedback diversity. OmniAD \cite{omniAD} relied mainly on supervised tuning, underutilizing GRPO’s exploratory strengths. These limitations highlight the need for GRPO adaptations tailored to AD’s unique constraints.
\section{Methodology}
\subsection{Revisiting GRPO}

Group Relative Policy Optimization (GRPO) is a post-training algorithm that aligns large language models (LLMs) with downstream tasks using reinforcement learning from preference feedback (RLPF).
Given a prompt \( x\), the policy model \( \pi_\theta \) generates multiple responses \( \{y_i\}_{i=1}^G \sim \pi_\theta(\cdot \mid x) \). Each response is assigned a scalar reward \( r(x, y_i) \in \mathbb{R} \), reflecting its quality under human or learned preferences. The model parameters \( \theta \) are updated via policy gradient to maximize estimated advantages (i.e., the normalized rewards):
\begin{equation}
\begin{split}
& \nabla_\theta J(\theta) = \mathbb{E}_{x, y_i} \left[ \hat{A}(x, y_i) \nabla_\theta \log \pi_\theta(y_i \mid x) \right],\\
& s.t.\quad \hat{A}(x, y_i) =\tilde{r} = \frac{r(x, y_i) - mean(r(x))}{std(r(x))}.
\end{split}
\end{equation}

This objective encourages the model to assign higher probabilities to high-reward responses while down-weighting less favorable ones. 

\paragraph{Reward Composition in GRPO.}
GRPO typically uses a reward function composed of multiple task-specific components. In AD, two main factors are considered:
\begin{equation}
r(x, y) = r_{\text{acc}}(x, y) + r_{\text{format}}(x, y),
\end{equation}
where \( r_{\text{acc}} \) measures response correctness against ground truth, and \( r_{\text{format}} \) assesses compliance with expected output structures (e.g., \textit{\textless think\textgreater \textless/think\textgreater \textless answer\textgreater \textless/answer\textgreater}).

\paragraph{GRPO with KL Regularization.}
To stabilize learning, GRPO augments the advantage-weighted log-likelihood with a KL divergence term between the current and reference policies:
\begin{equation}
\resizebox{0.9\linewidth}{!}{
$
\begin{aligned}
\mathcal{L}_{rew} &= - \mathbb{E}_{x} \frac{1}{G}\sum_{i=1}^G \frac{1}{|y_i|} \sum_{t=1}^{|y_i|} \left[\hat{A}(x, y_i) \log \pi_\theta(y_{i,t} \mid x) \right], \\
\mathcal{L}_{reg} &= \mathbb{D}_{KL}(\pi_\theta \parallel \pi_{\text{ref}}), \quad
\mathcal{L}(\theta) = \mathcal{L}_{rew} + \beta \cdot \mathcal{L}_{reg}
\end{aligned}
$
}
\end{equation}

where \( \pi_{\text{ref}} \) is a fixed baseline policy, e.g. the base model before finetuning, and \( \beta \) balances exploration and stability. The KL term prevents overly aggressive policy shifts.

\paragraph{Discussions in Anomaly Detection Task.}
The effectiveness of GRPO relies on two factors: the MLLM's ability to generate diverse, high-quality responses, and an effective scoring mechanism. The diversity in generated responses refers to the variance of the rewards assigned to each response. However, in anomaly detection, the task's simplicity, i.e. binary classification with only two possible outcomes, limits response diversity. To address this, we propose a Multi-Stage Reasoning architecture, which compels the MLLM to follow a structured reasoning process before generating an answer, thus avoiding making hasty, superficial answers and introducing more fine-grained reward signals. Furthermore, conventional, coarse-grained rewards, e.g., binary accuracy, can cause the learning signal to vanish when all responses share the same final answer, i.e. $\mathcal{L}_{rew}=0$. This not only limits sample utilization but also prevents distinguishing between genuinely and spuriously correct responses. To counter this, we analyze the reward variance and propose a Fine-Grained Reward method, which assesses both decision accuracy and localization quality within the reasoning process.

\subsection{Multi-Stage Reasoning}\label{sec:multi-stage}
To enhance the model’s reasoning capability and mitigate hasty decision-making in complex anomaly scenarios, we propose a structured, two-stage reasoning framework inspired by the test-time scaling paradigm and chain-of-thought prompting~\cite{madaan2023self, kojima2022large,wei2022chain}. This approach shifts from single-pass inference to a more deliberate, resource-adaptive process.
Recent progress in test-time scaling~\cite{madaan2023self} shows that allocating additional computation at inference can significantly improve performance in challenging cases. 
 Following these insights, our framework is designed as:

\begin{compactitem}
    \item \textbf{Stage 1: Initial Spatial Analysis (Think).} The model conducts a rapid, coarse-grained scan to localize potentially anomalous regions, generating region proposals for further scrutiny. This step also supplies the localization cues used by our reward mechanism.
    \item \textbf{Stage 2: Focused Region Examination (Rethink).} For each proposed region, the model allocates more reasoning capacity, enabling fine-grained analysis of suspicious areas to better distinguish true anomalies from benign variations.
    \item \textbf{Stage 3: Integrated Decision Making (Answer).} The model combines global and local information to make a final decision.
\end{compactitem}

    


This procedure aligns with the cognitive strategies employed by human experts in visual inspection tasks, 1) performing rapid global scanning to identify candidate regions of interest, 2) engaging in deliberate, focused analysis to distinguish genuine anomalies from normal variations. 

Additionally, from the GRPO training perspective, this deliberative, multi-stage reasoning framework provides a richer set of supervision signals, which facilitates more fine-grained reward assignment (as detailed in the next section) and offers a crucial mechanism for identifying responses that are spuriously correct. Furthermore, this framework enhances the diversity of the response space, thereby enriching the samples for more effective policy gradient estimation. 

\subsection{Fine-Grained Reward}

To address the limitations of binary rewards in GRPO-based anomaly detection, we propose a novel fine-grained reward mechanism. This mechanism fundamentally shifts the reward signal from a discrete, binary value to a more continuous, multi-task score by assessing both final decision accuracy and the explicit localization of anomalous regions.

\begin{figure}[!h]
    \centering
    \includegraphics[width=\linewidth]{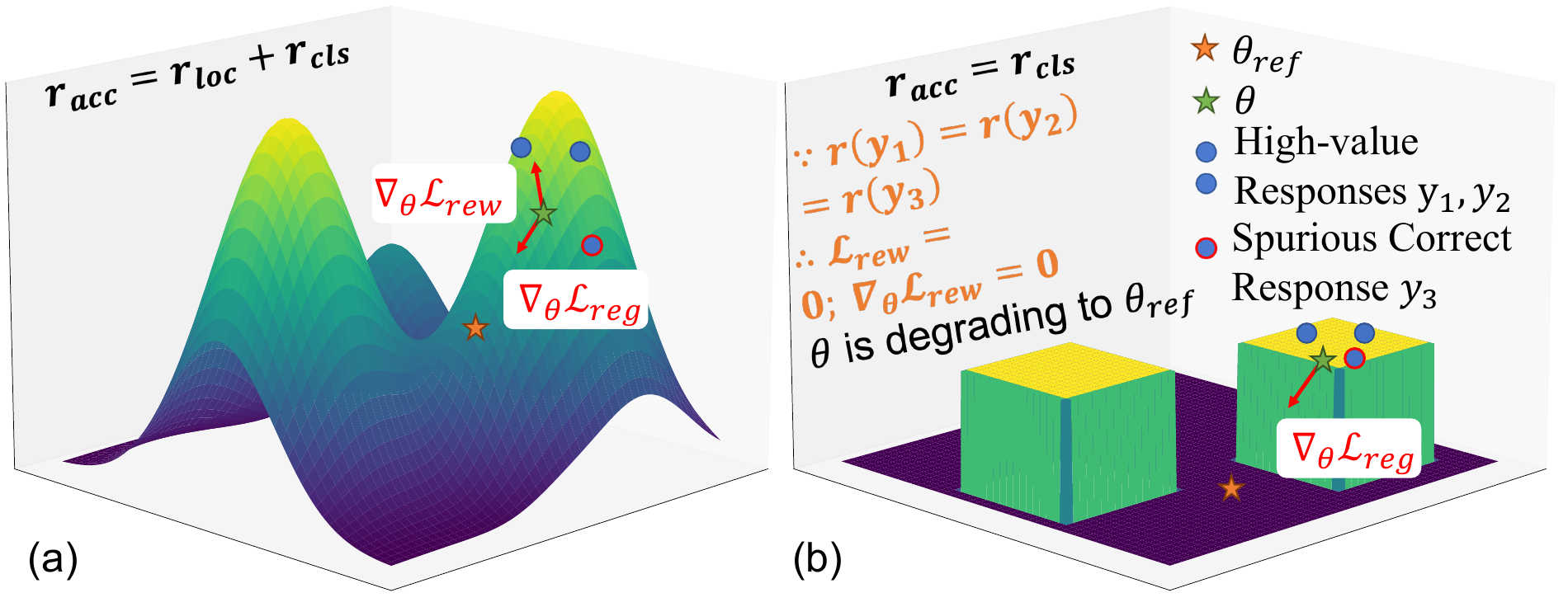}
    \caption{Conceptual illustration of loss landscape. (a) Our proposed fine-grained reward; (b) existing discrete reward. }
    \label{fig:reward}
\end{figure}

As shown in Fig. \ref{fig:reward},  traditional binary reward schemes (b) create flat, discontinuous optimization surfaces where all correct responses receive identical rewards, resulting in zero gradients i.e. $\nabla_\theta \mathcal{L}_{rew} = 0$ and model degradation toward the reference policy $\theta_{ref}$. In contrast, our localization-aware rewards (a) generate a rich, continuous landscape with meaningful gradients that provide strong directional signals: positive reinforcement for high-value responses $y_1,y_2$ with better spatial reasoning, and negative feedback for spurious correct responses $y_3$ that achieve classification accuracy through flawed inference.

To this end, two variants of fine-grained rewards oriented towards localization are introduced. Specifically, given a generated response $y$, we first extract the predicted bounding boxes $\mathcal{B}(y)=\{\hat{b}_i\}_{i=1}^{m}$ of anomalous regions from the response. The corresponding ground-truth bounding boxes in this sample denotes $\mathcal{B}(x) = \{b_j\}_{j=1}^{n}$. To assess the quality of the predicted anomalous regions, we use the average normalized GIoU over the matched box pairs, together with a count-based metric, as fine-grained reward signals. We define a cost matrix $C_{ij} = 1 - \mathrm{GIoU}(\hat{b}_i, b_j)$ and apply the Hungarian algorithm to find optimal matches \(\mathcal{M}(\mathcal{B}(y), \mathcal{B}(x)) \subseteq \{(i,j)\}_{i=1\dots m,\ \ j=1\dots n}\). The classification ground truth of this sample is indicated as $l(x) \in \{0, 1\}$ where $0$ denotes a normal sample while $1$ denotes an abnormal sample. We compute the following quantities,
\begin{equation}
\resizebox{\linewidth}{!}{$
r_{\text{loc}}(x, y) = 
\begin{cases}
\frac{1}{|\mathcal{M}|} \sum\limits_{(i,j) \in \mathcal{M}} \mathrm{GIoU}(\hat{b}_i, b_j) + \alpha \cdot r_{\text{count}}, & l(x) = 1, \\
\beta \cdot r_{\text{focus}}(m),\quad \text{s.t. } m = |\mathcal{B}(y)|, & l(x) = 0,
\end{cases}
$}
\end{equation}
with counting reward:
\begin{equation}
r_{\text{count}}(x, y) =
\begin{cases}
1, & |m - n| = 0 \\
0.5, & |m - n| = 1 \\
0, & |m - n| \geq 2
\end{cases},\ s.t.\ 
\begin{array}{l}
m = |\mathcal{B}(y)| \\
n = |\mathcal{B}(x)|.
\end{array}
\end{equation}
For normal samples, i.e. $l = 0$, where ground-truth boxes are absent, we still encourage the model to engage in focused analysis by incentivizing bounding box predictions via:
\begin{equation}
r_{\text{focus}}(m) =
\begin{cases}
0, & m = 0\\
0.5, & m = 1 \\
-0.1, & m \geq 2
\end{cases}
\end{equation}
This gently guides the model toward identifying a single region of interest, akin to human attention.
To ensure basic decision correctness, we include a classification reward, $r_{\text{cls}}(x, y) =\mathds{1}(l(y) = l(x))$, where $l(y)$ indicates the final classification answer. The overall accuracy reward integrates both localization and classification signals:
\begin{equation}
r_{\text{acc}}(x, y) = \omega \cdot r_{\text{loc}}(x, y) + r_{\text{cls}}(x, y).
\end{equation}

By providing this richer and more differentiated fine-grained reward signal, our method can effectively improve sample utilization and mitigate the model degradation situation while facilitate more accurate visual understanding.

\section{Experiments}
\label{others}
\subsection{Experiment Settings}

\paragraph{Datasets \& Protocols.}
To validate the adaptability of our method to diverse anomaly downstream tasks, we first follow a \textbf{multi-domain evaluation protocol} using four datasets that cover distinct notions of anomalies: MVTec~\cite{bergmann2019mvtec}, VisA~~\cite{zou2022spot}, MVTec-LOCO~\cite{bergmann2022beyond}, and GoodsAD~\cite{zhang2024pku}. This experimental design follows the benchmark protocol introduced in MMAD~\cite{jiang2025mmadcomprehensivebenchmarkmultimodal} and specifically evaluates the model’s ability to learn unified representations across heterogeneous anomaly types. For training, we use 20\% of the data, uniformly sampled across object categories, and reserve the remaining 80\% for in-distribution evaluation.

To further assess the practical utility of our approach in real-world industrial deployment, we evaluate on \textbf{cross-dataset generalization} setting. In this setting, the model is trained exclusively on 20\% of the MVTec dataset and tested on several unseen industrial anomaly detection benchmarks: VisA, MPDD~~\cite{MPDD}, DTD~~\cite{DTD}, and DAGM~~\cite{DAGM}. Each evaluation is conducted on the standard test split defined by the respective dataset. This setting is designed to simulate a common real-world constraint, where labeled training data is available only in a limited source domain, and models must generalize to novel types of anomalies in unseen target domains without retraining.

\paragraph{Competing Methods.}
The open-source models include LLaVA-Next~\cite{llavanext}, InternVL2~~\cite{InternVL2}, and InternVL3~~\cite{InternVL3}, while the proprietary baselines comprise GPT-4o, GPT-4o-mini, and GPT-4.1. In the \textbf{multi-domain evaluation} experiment, we additionally benchmark our approach against two concurrent MLLM fine-tuning methods: AnomalyR1~~\cite{AnomalyR1} and OmniAD~~\cite{omniAD}. 
For the \textbf{cross-dataset generalization} task, we include comparisons with state-of-the-art CLIP-based approaches~\cite{winclip, anomalyclip}. As CLIP-based methods naturally produce continuous anomaly scores, we convert these into binary decisions using Youden’s J statistic~\cite{youdensJ} to optimize the threshold $\tau$. Details of this method are provided in the supplementary. Additionally, we evaluate CLIP-based methods using a fixed threshold of $\tau=0.5$, representing a baseline approach that requires no dataset-specific optimization.



\paragraph{Implementation Details.}
All experiments were conducted on two NVIDIA B200 GPUs. During GRPO optimization, we generate six parallel responses per training sample, and train the model for 15 epochs. We adopt Low-Rank Adaptation (LoRA)~\cite{lora} to fine-tune the Qwen2.5-VL (7B) model~\cite{qwen25}, with LoRA hyperparameters rank $r=64$ and alpha $\alpha=128$. Due to hardware compatibility issues with FlashAttention on the Blackwell architecture, we instead employ standard attention mechanisms for all models. Importantly, only the LoRA parameters are updated during training, while the pre-trained backbone weights are frozen.


\subsubsection{Bounding Box Generation.}
Since the original datasets provide only pixel-level segmentation masks, we construct pseudo bounding boxes for localization supervision. Details are provided in the supplementary material.

\begin{table}[!htb]
 \centering
   \setlength{\tabcolsep}{2pt}
\resizebox{0.99\linewidth}{!}{
\renewcommand{\arraystretch}{0.95}
   \begin{tabular}{l|l|c|c|c|c|c|c}
   \toprule
         &  & \bf Scale & \multicolumn{1}{c|}{\bf MVTec} & \multicolumn{1}{c|}{\bf VisA} & \multicolumn{1}{c|}{\bf LOCO} & \multicolumn{1}{c|}{\bf GoodsAD} & \multicolumn{1}{c}{\bf Avg.} \\
         \midrule
   \multirow{3}{*}{\bf Proprietary} & GPT-4o-mini$^{\dagger}$ & -     & -      &  -     &  -     & -      & 64.33  \\
    & GPT-4o$^{\dagger}$ & -     &  -   &    - &  -   &  -   & 68.63 \\
    & GPT-4.1 & -     &   78.83 & \bf 80.99 &	\bf 69.35   & \bf 57.21  &  \underline{71.60} \\
   \midrule
   \multirow{5}{*}{\bf Open-source} & LLaVA-NeXT & 34B   & 72.44 & 55.91	& 63.28	& 56.06	& 61.92 \\
   & InternVL2 & 76B   & \underline{79.42} &	69.78 &	54.02 &	53.20 &	64.11 \\
    & InternVL3 & 8B & 78.53	& 65.02	& 59.77	& 50.20	& 63.38 \\
    & InternVL3 & 38B & 78.12	& 68.75	& 63.86	& 55.49	& 66.56 \\
    & QwenVL2.5 & 7B    & 62.03 & 55.50  & 45.90  & 40.73 & 51.04 \\
   \midrule
   \multirow{3}{*}{\bf Fine-tuned} & AnomalyR1$^{\ddag}$ & 3B    & - & - & - & - & 60.62 \\
    & OmniAD$^{\ddag}$ & 7B    & - & - & - & - & 68.80 \\
    & \bf AD-FM (Ours)  & 7B  &  \bf 90.72 & \underline{79.52} & \underline{65.62} & \underline{56.74} & \bf 73.15 \\
   \bottomrule
   \end{tabular}%
   }
   \caption{Binary classification accuracy (\%) for multi-domain anomaly detection. LOCO denotes MVTec-LOCO. Scale is measured in billions of parameters. $^{\dagger}$Results are from \protect\cite{jiang2025mmadcomprehensivebenchmarkmultimodal}. $^{\ddag}$Results are from respective publications.}
 \label{tab:id}%
\end{table}%

\subsection{Multi-Domain Anomaly Detection}

We first present the multi-domain anomaly detection results in Tab. \ref{tab:id}. Compared to the original Qwen model, our fine-tuned approach improves performance by 22.11\%. Our method also outperforms all evaluated large language models, including the latest commercial model GPT-4.1.


Against concurrent fine-tuning methods AnomalyR1 and OmniAD, our approach shows consistent superiority. Since the official implementations and detailed per-dataset results of both methods are unavailable, we directly report their average results as presented in their original papers. Our average performance exceeds AnomalyR1 by 12.53\% and OmniAD by 4.35\%. These improvements indicate that our design achieves better visual understanding within the GRPO fine-tuning paradigm.


%

We further conduct comprehensive evaluation across multiple anomaly analysis subtasks~\cite{jiang2025mmadcomprehensivebenchmarkmultimodal} to investigate the ability for anomaly explanation. 
As demonstrated in Tab.~\ref{tab:explain}, our approach consistently outperforms all baseline methods across the evaluated task categories. Detailed experimental protocols of these multi-task evaluations are provided in the supplementary materials.

\begin{table}[!htbp]
\centering
  \setlength{\tabcolsep}{2pt}
\resizebox{0.99\linewidth}{!}
{
\renewcommand{\arraystretch}{0.95}
\begin{tabular}{l|l|c|c|c|c|c|c}
\toprule
& & \bf Scale & \textbf{VisA} & \textbf{MPDD} & \textbf{DTD} & \textbf{DAGM} & \textbf{Avg.} \\
\midrule
\multirow{4}{*}{\textbf{CLIP-based}} & 
WinClip ($\tau = 0.5$)  & - & 43.40 &	\underline{72.10} &	25.90 & 86.90 & 57.08 \\
& WinClip* & - & 77.70 &	41.80 & 90.00 & 88.90 & 75.85 \\
& AnomalyClip ($\tau = 0.5$) & - & 56.56 & 56.49 & 81.71 & 22.19	& 54.24  \\
& AnomalyClip* & - & \bf 82.17 & 68.23 & 88.91 & \underline{95.36} & \underline{83.67} \\
\midrule
\multirow{3}{*}{\textbf{Proprietary}} & 
GPT4o-mini & -& 73.98 &	67.90	& 90.56	& 90.80	& 80.81 \\
& GPT4o & -& 73.77 &	69.43 &	\underline{90.80} & 90.17	 & 81.04 \\
& GPT-4.1 & -& \underline{80.99} &	66.81 &	90.18 &	89.93 &	81.98 \\
\midrule
\multirow{5}{*}{\textbf{Open-source}} 
& LLaVA-NeXT & 34B   & 55.91	& 61.57	& 72.62	& 23.15 & 53.70 \\
& InternVL2 & 76B   & 69.78	& 66.38	& 90.58	& 94.52	& 80.32 \\
& InternVL3 & 8B & 65.02	& 70.74	& 85.05	& 81.33	& 75.54 \\
& InternVL3 & 38B & 68.75	& 69.43	& 90.79	& 89.87	& 79.71 \\
& Qwen2.5-VL & 7B & 55.50 & 68.34 & 85.59 & 94.83 & 76.07 \\
\midrule
\bf Fine-tuned & \bf AD-FM (Ours) & 7B & 77.43 & \bf 72.71 & \bf 92.64 & \bf 95.51 & \bf 84.57 \\
\bottomrule
\end{tabular}
}
\caption{Cross-dataset generalization performance in industrial anomaly detection. Binary classification accuracy (\%) with single normal reference protocol. *denotes threshold optimized using Youden's J statistic.}
\label{tab:ood}
\end{table}

\subsection{Cross-Dataset Generalization in Industrial Anomaly Detection}

The cross-dataset generalization results on industrial anomaly detection tasks are presented in Tab.~\ref{tab:ood}, where all reported accuracies are obtained using our model trained exclusively on the MVTec dataset. Without any additional training on the target datasets, our method achieves a notable average improvement of 8.5\% over the base Qwen2.5-VL (7B) model. It also consistently outperforms all evaluated zero-shot MLLM baselines. Compared to CLIP-based methods, our approach surpasses them on three out of four datasets (MPDD, DTD, and DAGM), while avoiding the critical reliance on threshold tuning.

On the VisA dataset, where our model shows a modest gap relative to the optimized CLIP baseline, the shortfall primarily arises from the dataset’s subtle defect patterns, which pose significant challenges for current MLLMs. Nonetheless, our method offers substantial practical benefits. CLIP-based methods are highly sensitive to threshold settings, e.g. AnomalyCLIP's performance drops from 83.67\% (optimized) to 54.24\% (fixed threshold), and WinCLIP from 75.85\% to 57.08\%. In contrast, our method provides direct binary predictions without requiring threshold calibration, which is an essential advantage for real-world deployments where full data distributions are not available for tuning.

\begin{table*}[!htb]
\centering
   \setlength{\tabcolsep}{3pt}
\resizebox{0.9\linewidth}{!}{
\renewcommand{\arraystretch}{0.85}
\begin{tabular}{l|l|c|c|c|c|c|c|c|c}
\toprule
& & \textbf{Scale} & \textbf{Defect Class.} & \textbf{Defect Loc.} & \textbf{Defect Desc.} & \textbf{Defect Anal.} & \textbf{Object Class.} & \textbf{Object Anal.} & \textbf{Avg.} \\
\midrule
\multirow{3}{*}{\textbf{Proprietary}} & GPT-4o-mini$^{\dagger}$ & - & 48.58 & 38.75 & 63.68 & 80.40 & 88.56 & 79.74 & 66.62 \\
& GPT-4o$^{\dagger}$ & - &  65.80 & 55.62 & 73.21 & 83.41 & \underline{94.98} & 82.80 & 75.97 \\
& GPT-4.1 & - & 65.93 &  62.64 &  76.18 &  84.23  & 90.95  & 85.57  & 77.58 \\
\midrule
\multirow{5}{*}{\textbf{Open-source}} 
& LLaVA-NeXT$^{\dagger}$ & 34B & 48.79	& 52.87	& 71.34	& 80.28 & 	81.12	& 77.80	& 68.70 \\
& InternVL2$^{\dagger}$ & 76B & 54.22 & 56.66 & 66.30 & 80.47 & 86.40 & 82.92 & 71.16 \\
& InternVL3 & 8B & 46.35	& 46.24	& 66.01	& 78.28 & 79.17	& 84.83 & 66.81 \\
& InternVL3 & 38B & 58.99	& 59.63	& 73.06	& 83.88	& 91.62	& 88.50	& 75.94 \\
& Qwen2.5-VL & 7B & 41.07 & 44.87 & 61.43 & 75.99 & 86.54 & 79.58 & 64.91 \\
\midrule
\multirow{3}{*}{\textbf{Fine-tuned}} & AnomalyR1$^{\ddag}$ & 3B & 63.56 & 70.14 & \underline{80.47} & 85.28 & 92.48 & 86.15 & 79.68 \\
& OmniAD$^{\ddag}$ & 7B & \bf 78.80 & \underline{75.50} & 67.20 & \underline{86.40} & \bf 96.00 & \underline{86.40} & \underline{81.72} \\
& \bf AD-FM (Ours) & 7B & \underline{73.36} & \bf 77.53 & \bf 86.80 & \bf 86.75 & 89.98 & \bf 86.67 &  \bf 83.56 \\
\bottomrule
\end{tabular}
\renewcommand{\arraystretch}{1.0}
}
\caption{Multi-task anomaly analysis performance comparison across six evaluation categories. Results reported as average accuracy (\%) across four datasets (MVTec, VisA, MVTec-LOCO, GoodsAD) for defect-related tasks and object-level understanding tasks. $^{\dagger}$Results are from \protect\cite{jiang2025mmadcomprehensivebenchmarkmultimodal}. $^{\ddag}$Results are from respective publications.}
\label{tab:explain}
\end{table*}

\subsection{Qualitative Results}

\begin{figure*}[htbp]
    \centering
    \includegraphics[width=\textwidth]{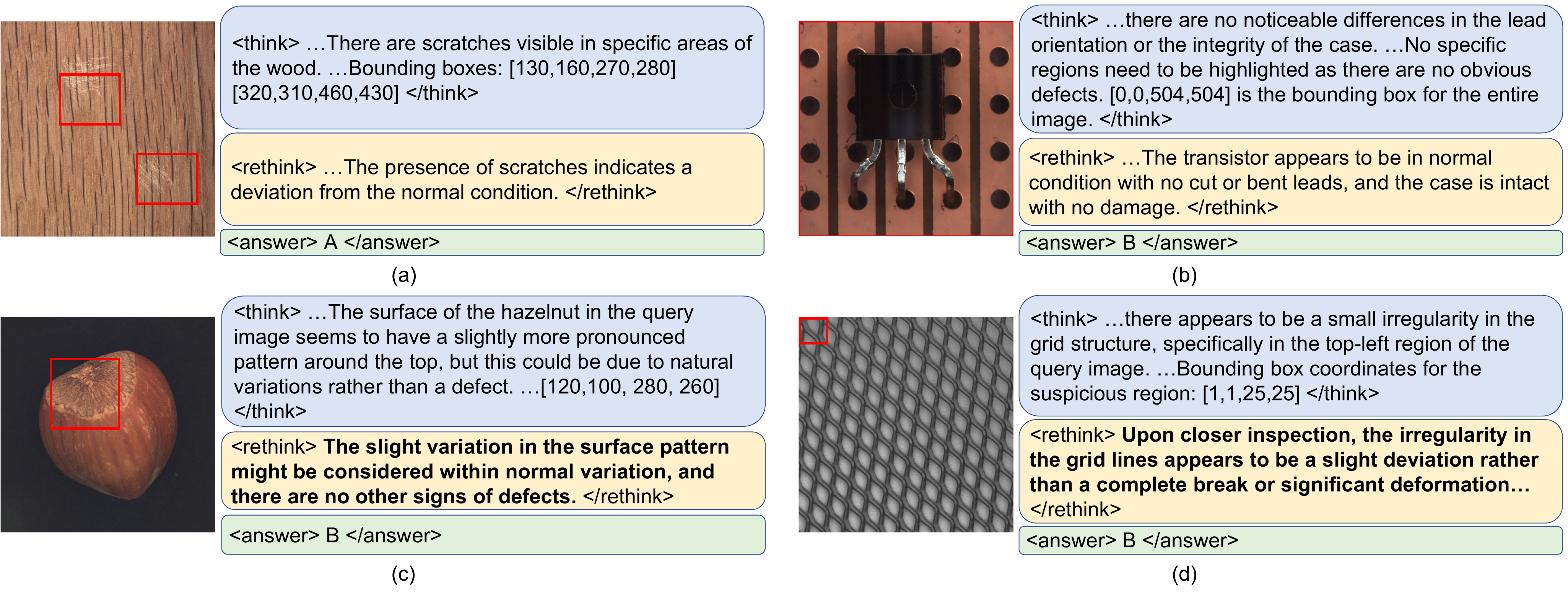}
    \caption{Qualitative examples demonstrating the effectiveness of our multi-stage deliberative reasoning framework.}
    \label{fig:samples}
\end{figure*}

\paragraph{Multi-stage Deliberative Reasoning.}
To demonstrate our multi-stage deliberative reasoning framework, we present representative examples in Fig. \ref{fig:samples} that highlight different aspects of our approach's reasoning capabilities.

The first two examples demonstrate successful detection of obvious anomalies and accurate classification. In case (a), the model correctly identifies surface irregularities on wood in \textit{think} stage, noting deviations from natural grain patterns with appropriate spatial localization. Case (b) illustrates how our framework cultivates systematic analytical habits even for normal samples while no ground truth bounding boxes are available for normal cases. 

Cases (c) and (d) illustrate the critical value of our multi-stage approach in handling initially ambiguous scenarios.
In case (c), during \textit{think} stage, the model exhibits uncertainty, noting that \textit{``the slight variation in the surface pattern might be considered within normal variation, and there are no other signs of defects.''}. However, the \textit{rethink} stage enables deeper examination, ultimately leading to correct normal decision. Similarly, case (d) progresses from identifying \textit{``a small irregularity in the grid structure,''} to determining through \textit{``closer inspection''} that the deviation constitutes \textit{``a complete break or significant deformation.''}

This progression from uncertainty to confident demonstrates how our framework prevents premature conclusions and enables recovery from initial analytical uncertainty.

\begin{figure}[!htbp]
    \centering
    \includegraphics[width=0.99\linewidth]{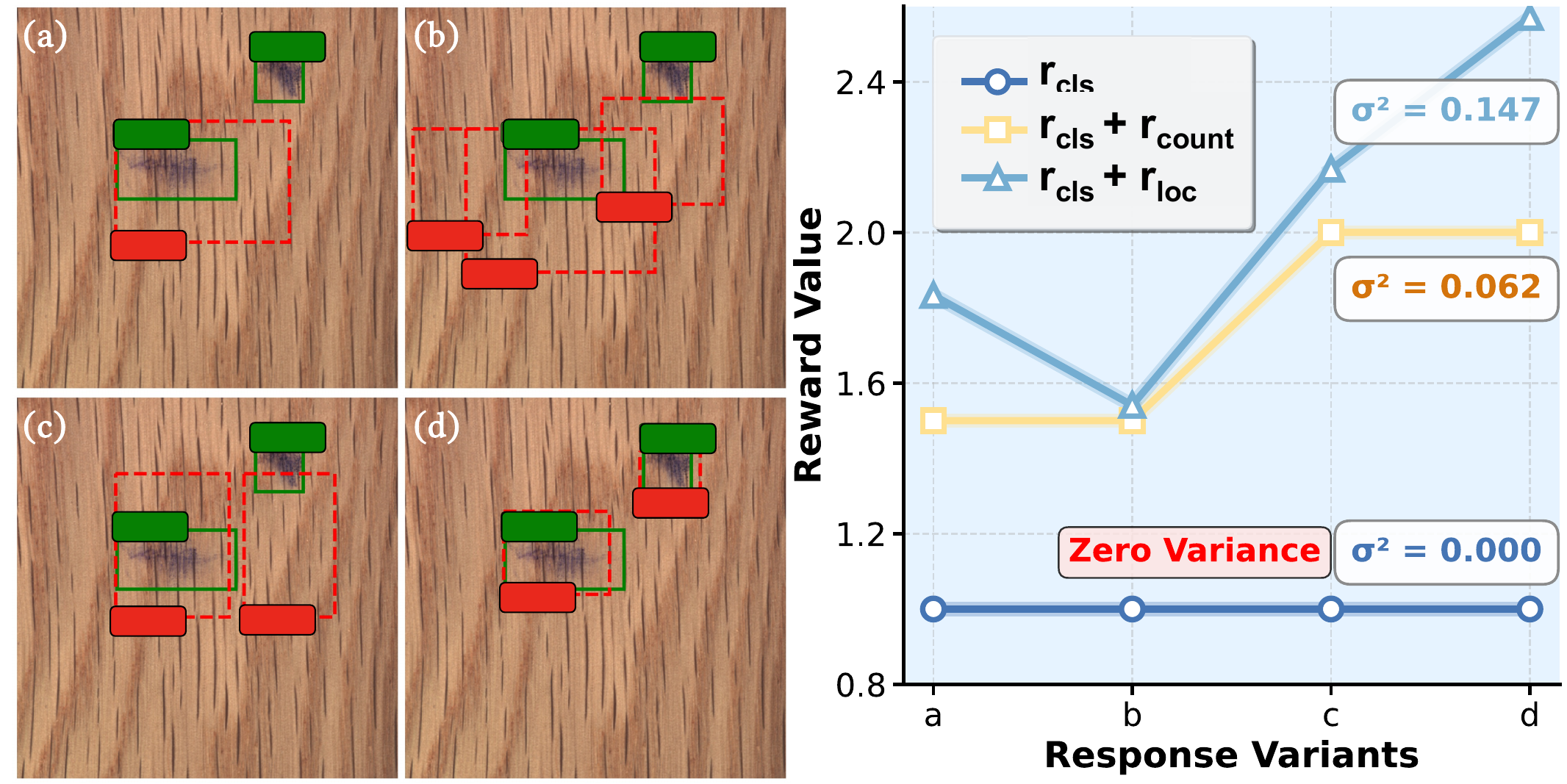}
    \caption{Response variance visualization for the same training sample under different reward mechanisms. Left: Four model responses (a-d) showing varying quality in defect localization and spatial reasoning. Right: Reward variance analysis under different reward schemes.}
    \label{fig:responses_samples}
\end{figure}

\paragraph{Response Variance Visualization.}
To provide deeper insight into how our fine-grained rewards differentiate responses with greater sophistication, we visualize the responses generated by our framework on a representative sample. Fig. \ref{fig:responses_samples} presents four different responses (4 responses chosen from all 6 due to space limitations) generated by our model for the same input image containing wood surface defects, alongside the corresponding reward variance analysis.

As seen from Fig.~\ref{fig:responses_samples} (left), while all responses (a-d) correctly identify anomaly presence, there is a clear progression in analytical quality from (a) to (d). Specifically, responses (a-b) demonstrate incorrect defect enumeration and imprecise localization, response (c) shows improved awareness but lacks precision, and response (d) achieves accurate counting with precise bounding box alignment. 

Further examining the reward value plot in Fig.~\ref{fig:responses_samples} (right), under binary reward schemes, i.e., $r_{\text{cls}}$, all four responses receive identical scores since they all correctly identify anomaly presence, resulting in zero variance and no learning signal for GRPO optimization. 
In contrast, our fine-grained reward mechanisms inject meaningful variance into the training signal. The count-based enhancement, i.e., $r_{\text{cls}} + r_{\text{count}}$, generates moderate variance while failing to differentiate between response (c) and (d), while our localization-aware formulation, i.e., $r_{\text{cls}} +r_{\text{loc}}$, provides  feedback that incentivizes the most accurate response (d) over the progressively less precise responses. This granular differentiation enables GRPO to learn from the full spectrum of analytical quality rather than treating all correct detections as equivalent. This further improves the interpretability and reliability of the reasoning process, contributing to safer and more transparent model behavior.

\subsection{Ablation Studies}
\paragraph{Component Analysis.}
\begin{table}[htbp]
  \centering
     \setlength{\tabcolsep}{2pt}
  \resizebox{0.99\linewidth}{!}{
  \renewcommand{\arraystretch}{0.95}
    \begin{tabular}{l|c|c|c|c|c|c}
    \toprule
    \bf Reward $r_{\text{acc}}$ & \bf Reasoning &  \bf MVTec & \bf VisA & \bf LOCO & \bf GoodsAD & \bf Avg. \\
    \midrule
    -     &  -     & 62.03 & 55.50  & 45.90  & 40.73 & 51.04 \\
    $r_{\text{cls}}$ & one-stage & 84.64 & 73.79	& 60.08	& 53.77	& 68.07 \\
    $r_{\text{cls}}$ & multi-stage & 86.67	& 75.29	& 61.42	& 55.91	& 69.82 \\
    $r_{\text{cls}} + r_{\text{random}}$ & multi-stage &  88.41	& 75.44	& 60.60	& 56.44	& 70.22 \\
    $r_{\text{cls}} + r_{\text{loc}}$ & multi-stage & \bf 90.72 & \bf 79.52 & \bf 65.62 & \bf 56.74 & \bf 73.15 \\
    \bottomrule
    \end{tabular}%
    }
  \caption{Ablation study on reward mechanisms and reasoning processes on multi-domain AD experiments. Performance measured as binary classification accuracy (\%). }
  \label{tab:ablation}%
\end{table}%

To validate the effectiveness of our GRPO optimization components, we conduct comprehensive ablation studies across the four multi-domain dataset, systematically examining the contributions of different reward mechanisms and reasoning protocols. The results are demonstrated in Tab.~\ref{tab:ablation}. From the results, we make the following key observations:

\begin{compactitem}
    \item The baseline configuration without any training achieves an average accuracy of 51.04\%, representing the inherent zero-shot capability of the base Qwen2.5-VL model.  The introduction of \textbf{binary reward signals with one-stage reasoning} boosts performance to 68.07\% (+17.03\%), demonstrating effective domain-specific fine-tuning.
    \item Extending the binary reward framework to incorporate \textbf{multi-stage reasoning} processes further enhances performance to 69.82\% (+1.75\% over one-stage), demonstrating that structured multi-step inference protocols enable more sophisticated anomaly analysis capabilities. 
    \item To isolate the contribution of variance enhancement from structured localization supervision, we evaluate a \textbf{random reward} component $r_{random}$ that introduces artificial variance without spatial meaning. This approach builds upon Random Reward Perturbation (RRP) \cite{randomreward}, where zero-mean Gaussian noise added to reward signals has been shown to enhance exploration through increased policy diversity. This configuration achieves 70.22\% with multi-stage reasoning, providing modest improvements (+0.40\% over structured reasoning alone), confirming that variance injection per se contributes positively but limited to GRPO effectiveness.
    \item In contrast, our \textbf{fine-grained reward} formulation achieves 73.15\% (+3.33\% over binary multi-stage), representing a cumulative improvement of +22.11\% over the zero-shot baseline. This performance gap between random and ours variance injection validates that our designed rewards deliver dual benefits: variance enhancement that provides essential signal diversity, and fine-grained quality assessment that selectively reinforces responses demonstrating genuine understanding while filtering out spurious correctness patterns.
\end{compactitem}



\paragraph{Response Variance Analysis.}
To empirically validate our hypothesis that continuous reward mechanisms enhance GRPO training effectiveness by increasing response variance, we conduct an ablation analysis on MVTec dataset. 
We uniformly sample 500 images from MVTec dataset and generate 6 responses per sample using Qwen2.5-VL model under different reward configurations. For each sample, we compute the variance in reward values across 6 generated responses. We then measure the percentage of samples with zero variance, indicating cases where all responses receive identical rewards and thus contribute no meaningful gradient signal to GRPO optimization. Fig. \ref{fig:var} presents the zero-variance percentages across different reward mechanisms:

\begin{figure}[htbp]
    \centering
    \includegraphics[width=\linewidth]{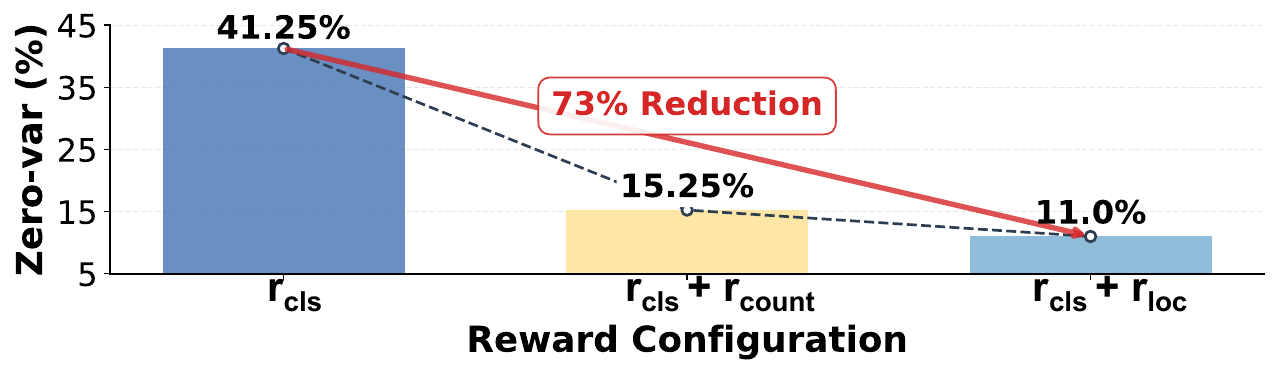}
    \caption{Percentage of samples with zero reward variance across 6 generated responses under different reward formulations. Lower percentages indicate enhanced data utilization for GRPO training.}
    \label{fig:var}
\end{figure}


The results demonstrate a significant improvement in data utilization through our reward design. The binary-only baseline yields 41.25\% zero-variance samples, which empirically confirms the challenge we identified: traditional reward schemes create sparse learning environments that severely limit gradient flow. In contrast, adding bounding box counting reduces this to 15.25\%, and our complete formulation with localization lowers it further to 11.00\%, which is a 73\% reduction compared to the binary baseline. These results empirically validate our theoretical motivation: continuous reward mechanisms transform GRPO from a sparse-signal optimization into a rich-feedback learning environment.

\section{Conclusion}
In this paper, we propose AD-FM, a fine-tuned multi-modal MLLM framework tailored for anomaly detection, addressing two core limitations of existing GRPO-based applications: response variance collapse and under-constrained reasoning procedure. Our multi-stage deliberative reasoning creates supervisory signals and promotes cautious inference, while our fine-grained, spatially grounded reward design introduces dense and discriminative learning signals that suppress spurious correctness. These components jointly enhance model interpretability, training efficiency, and generalization. Extensive results across multi-domain and cross-dataset evaluations demonstrate the effectiveness of our approach in adapting general-purpose MLLMs to anomaly detection tasks. 

\section{Acknowledgments}
This research is supported by Agency for Science, Technology and Research (A*STAR) under its MTC Programmatic Funds (Grant No. M23L7b0021); and the National Research Foundation, Singapore, and the CyberSG R\&D Programme Office (``CRPO''), under the National Cybersecurity R\&D Programme (``NCRP''), RIE2025 NCRP Funding Initiative (Award CRPO-GC1-NTU-002).

\bibliography{aaai2026}

\end{document}